\documentclass[11pt,a4paper]{article}
\usepackage[hyperref]{emnlp2020}
\usepackage{times}
\usepackage{latexsym}

\usepackage{microtype}
\usepackage{algorithm}

\aclfinalcopy 

\usepackage{balance}
\usepackage{booktabs}
\usepackage{times}
\usepackage{latexsym}
\usepackage{multirow}
\usepackage[utf8]{inputenc}
\usepackage{graphicx}
\usepackage{amsmath}
\usepackage{makecell}
\usepackage{algorithmic}
\usepackage{algorithm}

\usepackage{todonotes}
\presetkeys{todonotes}{inline}{}

\newcommand{\nlp}{\textsc{NLP}}
\newcommand{\eda}{\textsc{EDA}}
\newcommand{\bt}{\textsc{BT}}
\newcommand{\mnli}{\textsc{MNLI}}
\newcommand{\trec}{\textsc{TREC}}
\newcommand{\sst}{\textsc{SST-2}}
\newcommand{\sts}{\textsc{STS-B}}
\newcommand{\rt}{\textsc{RT}}
\newcommand{\subj}{\textsc{SUBJ}}
\newcommand{\bert}{\textsc{BERT}}
\newcommand{\xlnet}{\textsc{XLNet}}
\newcommand{\roberta}{\textsc{RoBERTa}}
\newcommand{\lstm}{\textsc{LSTM}}
\newcommand{\respace}{\vspace*{-.1cm}}

\title{How Effective is Task-Agnostic Data Augmentation for \\Pretrained Transformers?}

\newcommand*\samethanks[1][\value{footnote}]{\footnotemark[#1]}

\author{Shayne Longpre\thanks{\hspace{2mm}equal contribution} \\
  Apple Inc. \\
  \texttt{slongpre@apple.com} \\\And
  Yu Wang\samethanks \\
  Apple Inc. \\
  \texttt{w.y@apple.com} \\\And
  Christopher DuBois \\
  Apple Inc. \\
  \texttt{cdubois@apple.com} \\}

\date{}

\begin{document}
\maketitle
\begin{abstract}
Task-agnostic forms of data augmentation have proven widely effective in computer vision, even on pretrained models. 
In \nlp{} similar results are reported most commonly for low data regimes, non-pretrained models, or situationally for pretrained models. 
In this paper we ask how effective these techniques really are when applied to pretrained transformers. 
Using two popular varieties of task-agnostic data augmentation (not tailored to any particular task), Easy Data Augmentation \citep{wei2019eda} and Back-Translation \citep{sennrich2015improving}, we conduct a systematic examination of their effects across 5 classification tasks, 6 datasets, and 3 variants of modern pretrained transformers, including \bert{}, \xlnet{}, and \roberta{}. 
We observe a negative result, finding that techniques which previously reported strong improvements for non-pretrained models fail to consistently improve performance for pretrained transformers, even when training data is limited. 
We hope this empirical analysis helps inform practitioners where data augmentation techniques may confer improvements.
\end{abstract}

\respace
\section{Introduction}
\respace

``Task-agnostic" data augmentations --- those which are not tailored to a task, but are broadly applicable across the visual or textual domain --- have long been a staple of machine learning.
Task-agnostic data augmentation techniques for computer vision, such as image translation, rotation, shearing, and contrast jittering, have achieved considerable success, given their ease of use, and wide-spread applicability \citep{cubuk2018autoaugment, perez2017effectiveness}. 
In natural language processing, benefits of data augmentation have usually been observed where the augmentations are suited to the task: as with back-translation for machine translation \citep{edunov-etal-2018-understanding, xia2019generalized}, or negative sampling for question answering and document retrieval \citep{10.1145/3077136.3080645, yang2019data, xiong2020approximate}. 
Outside of application-tailored augmentations, improvements are primarily reported on autoregressive models without unsupervised pretraining or contextual embeddings, such as LSTMs and CNNs, and even then in low data regimes \citep{NIPS2015_5782, coulombe2018text, wei2019eda, yu2018qanet}. 
Additionally, in computer vision task-agnostic augmentations continue to report benefits when applied to pretrained representations \citep{gu2019improve}.
However, in \nlp{} it is less clear whether these general augmentations benefit modern Transformer architectures with unsupervised pretraining at scale.

We pose the question: to what extent do modern \nlp{} models benefit from task-agnostic data augmentations? 
In this paper, we provide empirical results across a variety of tasks, datasets, architectures, and popular augmentation strategies.
Among data augmentation techniques, we select Easy Data Augmentation \citep{wei2019eda} and Back-Translation \citep{sennrich2015improving}; \eda{} and \bt{} respectively.
Both are popular task-agnostic options, and report significant gains for LSTMs on a wide variety of datasets.
We apply these techniques to 6 classification-oriented datasets, spanning 5 tasks with varying linguistic objectives and complexity.
For fair comparison, we tune each of \bert{}, \xlnet{}, and \roberta{} extensively, allocating an equal budget of trial runs to models trained with and without augmentations.
As a separate dimension, we also vary the availability of training data to understand under what specific conditions data augmentation is beneficial. 

Our findings demonstrate that these popular task-agnostic data augmentations provide only sparse and inconsistent improvements for modern pretrained transformers on many simple classification tasks.
They further lend empirical evidence to the hypothesis that task-agnostic data augmentations may be significantly less effective on pretrained transformers for other classification and NLP tasks.
Observed patterns suggest that the scale of pretraining may be the critical factor replacing the need for linguistic variety that augmentations confer.
We hope our work provides guidance to ML practitioners in deciding when to use data augmentation and encourages further examination of its relationship to unsupervised pretraining.

\respace
\section{Experimental Methodology}
\respace
\label{sec:exp-setup}

\respace
\subsection{Datasets}
\respace

Following ~\citet{wei2019eda} and ~\citet{wu2019conditional} we adopt 4 classification datasets on which general data augmentation techniques demonstrated strong performance gains, and include 2 more from the \textsc{GLUE} benchmark \citep{wang2018glue}. 
As shown in Table~\ref{table:datasets}, a variety of classification sizes, sequence lengths, and vocabulary sizes are represented. 
Included tasks are sentiment analysis (\sst{}, \rt{}), subjectivity detection (\subj{}), question type classification (\trec{}), semantic similarity (\sts{}) and natural language inference (\mnli{}).

\begin{table}[t]
	\centering
	\resizebox{\columnwidth}{!}{
			\begin{tabular}{lrrrr}
				\toprule
				Dataset & $c$ & $l$ & $|D_{train}|$ \\
				\midrule
                {SST-2} \citep{socher2013recursive} & 2 & 19 & 7.6k \\
                {SUBJ} \citep{10.3115/1218955.1218990} & 2 & 23 & 8k \\
                {RT} \citep{pang2005seeing} & 2 & 21 & 8.7k \\
                {MNLI} \citep{williams2017broad} & 3 & 2x17 & 8k \\
                {STS-B} \citep{baudivs2016sentence} & 5 & 2x12 & 6.6k \\
                {TREC} \citep{10.3115/1072228.1072378} & 6 & 10 & 3.9k \\
				\bottomrule
			\end{tabular}}
	\caption{Summary statistics for each dataset. $c$: The number of classes. $l$: The average sequence length in word tokens. $D_{train}$: The training set size after random sampling up to $10k$ unique examples (if available), and subtracting 2k for dev and test sets.}
	\label{table:datasets}
	\vspace*{-.4cm}
\end{table}

\respace
\subsection{Augmentation Techniques}
\respace

Among the many variations of data augmentation two families are widely used in \nlp{}: back translation and text editing. 

\textbf{Back Translation (\bt{}):}  We use an English to German machine translation model \citep{ott2018scaling} and a German to English model \citep{ng2019facebook}.\footnote{Adapted from \url{https://ai.facebook.com/tools/fairseq/}.}
We selected German due to its strong results as a pairing with English for back translation, as reported in \citet{yu2018qanet, sennrich2015improving}.
We translate each English sentence to one German sentence and back to six candidate English sentences. From these sentence candidates we obtain the best results sampling the most distant sentence from the original English sentence, measured by word edit distance. From manual inspection this approach produced the most diverse paraphrases, though this strategy needs to be tailored to the machine translation systems employed.
The overall aim of this strategy is to maximize linguistic variety while retaining sentence coherency.

\textbf{Easy Data Augmentation (\eda{}):} Following \citet{wei2019eda} we employ a combination of popular text editing techniques that have shown strong performance on LSTMs.\footnote{We use the implementation at \url{https://github.com/jasonwei20/eda_nlp}.}
Text edits include synonym replacement, random swap, random insertion, and random deletion. 
To improve upon \eda{} further, we enforce part-of-speech consistency for synonym selection.
As an example, the verb ``back" in the phrase ``to back the government" will not be replaced by ``rear", which is a synonym of the noun ``back". 

\respace
\subsection{Experimental Setup}
\respace

To conduct a fair assessment of each data augmentation technique, we ensure three properties of our experimental setup: (I) our tuning procedure mimics that of a machine learning practitioner; (II) the selected hyperparameters cannot be significantly improved as to change our conclusions; and (III) each strategy is evaluated with an equal number of trial runs.\footnote{We verify property (II), that the tuning budget is sufficient, by doubling the allocated trials and observing the magnitude of changes (see Appendix B).}

We experiment with 3 types of Transformers \citep{vaswani2017attention}: \textsc{Bert-Base} \citep{devlin2019bert}, \textsc{XLNet-Base} \citep{yang2019xlnet}, and \textsc{RoBERTa-Base} \citep{liu2019roberta}.
These models each use slightly different pretraining strategies.
\bert{} and \roberta{} are both pretrained with Masked Language Modeling, but with different auxiliary objectives and number of training steps.
\xlnet{} is pretrained with its own ``Permutation Language Modeling".
For each model and dataset $1k$ examples are randomly selected for each of the validation and test sets. 
Separately from these fixed sets, we iterate over five training data sizes $N \in \{500, 1000, 2000, 3000, \text{Full}\}$ to simulate data scarcity. 

Given a model $M$, dataset $D$, and training set size $N$, we allocate an equal number of training runs to No Augmentation (\textsc{No DA}), \eda{}, and \bt{}. 
For each setting, we define continuous ranges for the learning rate, dropout, and number of epochs. 
\eda{} and \bt{} settings also tune a ``dosage" parameter governing augmentation $\tau \in \{0.5, 1, 1.5, 2\}$. 
$N\times\tau$ is the quantity of augmented examples added to the original training set. 
\begin{algorithm}[t]
    \caption{\textsc{Experimental Design}}
  \label{alg:experiment}
\begin{algorithmic}[1]
  \footnotesize
  \REQUIRE{Model $M$, Dataset $D$, Train size $N$}
  \ENSURE{Mean and standard deviation for test accuracies ($\mu_{NoDA}$, $\sigma_{NoDA}$), ($\mu_{BT}$, $\sigma_{BT}$), ($\mu_{EDA}$, $\sigma_{EDA}$)} \\
  \STATE $T_1,\ T_2, K$ $\gets$ $3,\ 20,\ 30$
  \STATE $D_{train}$ $\gets$ {sample(shuffle($D$), $N$)}
  
  \FOR{each Augmentation $\alpha$ in $[NoDA,BT,EDA]$}
  \STATE // \textit{Find best hyperparameters $H_{\alpha}$ for augmentation $\alpha$}\\
  \STATE $H_{\alpha} \leftarrow $ \textsc{RandomSearch}($M$, $D_{train}$, $K$, $T_1$)\\
  \STATE $M_{\alpha} \leftarrow $ $M.$\textsc{Use}($H_{\alpha}$) \\ \label{alg:apply-hyp}
  
  \STATE // \textit{Compute validation scores for augmentation $\alpha$}
  \FOR{$s = 1$ \TO $T_2$}
  \STATE \textsc{Scores} $\leftarrow$ \textsc{Train}($M_{\alpha}$, $D_{train}$, seed=$s$) \\
  \ENDFOR

  \STATE // \textit{Select test scores using best validation scores}
  \STATE $\mu_{\alpha}, \sigma_{\alpha} \leftarrow $ \textsc{SelectBest}(Scores, $10$) \\ \label{alg:select-best}
  \ENDFOR

  \RETURN ($\mu_{NoDA}$, $\sigma_{NoDA}$), ($\mu_{BT}$, $\sigma_{BT}$), ($\mu_{EDA}$, $\sigma_{EDA}$)
\end{algorithmic}
\end{algorithm}

First, we conduct a \textsc{RandomSearch} for $K=30$ parameter choices, each repeated for $T_1=3$ trials with differing random training seed.
As shown in Algorithm~\ref{alg:experiment} this stage returns the optimal hyperparameter choices $H_{\alpha}$ for each augmentation type $\alpha \in \{\text{NoDA},\ \text{BT},\ \text{EDA}\}$. 
The best hyperparameters are selected by mean validation accuracy over random seed trials. 
In the second stage, a model with these best hyperparameters (Algorithm~\ref{alg:experiment} line \ref{alg:apply-hyp}) is trained over $T_2=20$ random seed trials.\footnote{Note that we cache the top performing trials from $T_1$ to reduce total trial runs.} 
Finally, the $10$ best trials by validation accuracy are selected for each per setting (line \ref{alg:select-best}).
We report the mean and $95\%$ confidence intervals of their test results.
The bottom $10$ trials are discarded to account for the high accuracy variance of pretrained language models with respect to weight initialization, and data order \citep{dodge2020fine}.
This procedure closely mimics that of an ML practitioner looking to select the best model.\footnote{Further details for our model tuning procedure are available in Appendix A.}

\begin{figure*}
	\centering
	\includegraphics[width=\linewidth]{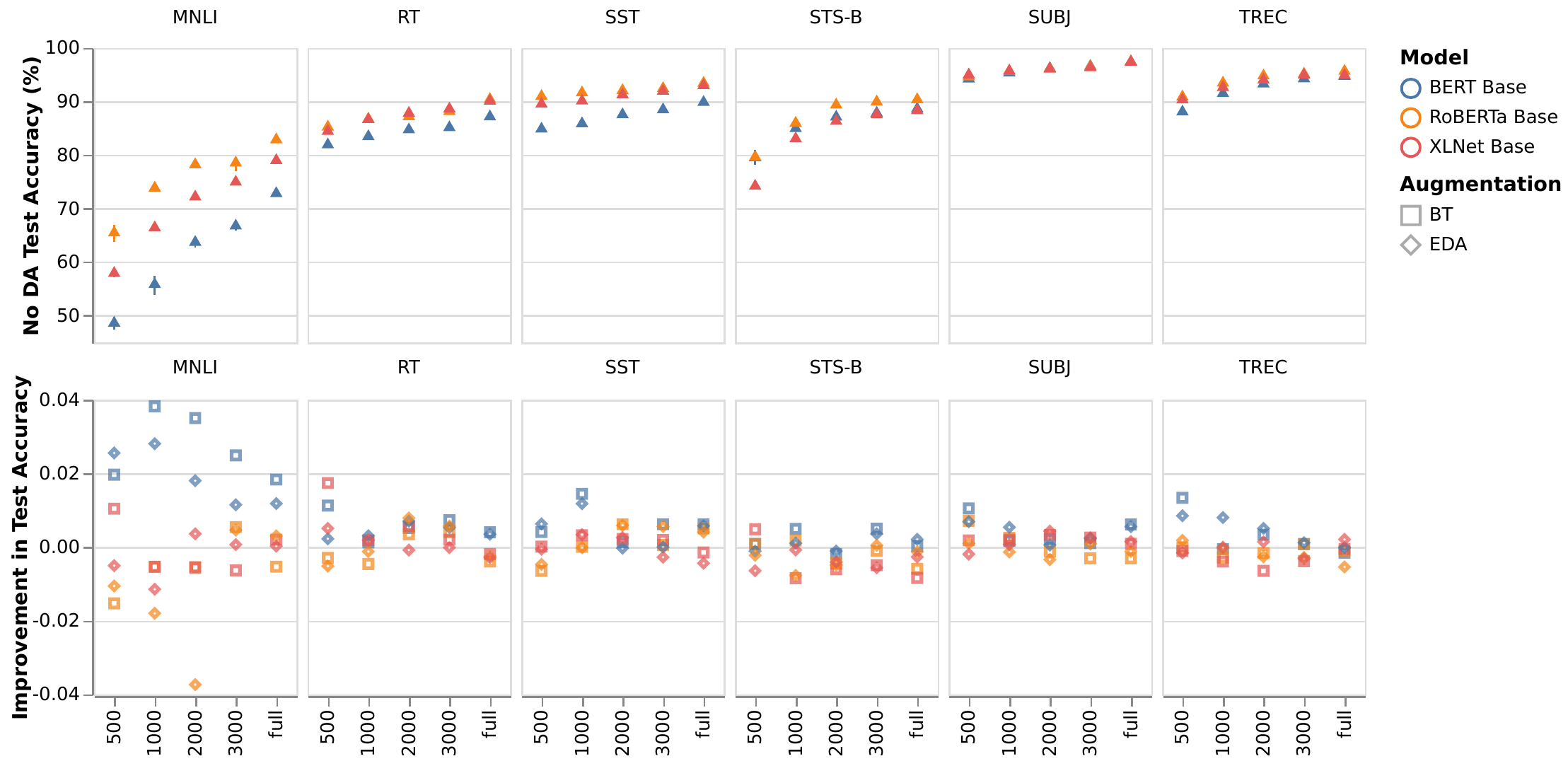}
	\caption{The upper row plots the mean test accuracies, with $95\%$ confidence intervals, across model types, datasets, and training sizes for the \textsc{No DA} baseline setting (displayed with triangle points). The lower row plots the mean relative improvement in test accuracy over the \textsc{No DA} setting, for each augmentation type.}
	\label{fig:full-results}
	\vspace*{-.4cm}
\end{figure*}

\respace
\section{Empirical Results}
\respace
\label{sec:results}
	  
Figure~\ref{fig:full-results} shows both the baseline \textsc{No DA} test accuracies as a reference point, and the mean relative improvement from applying \eda{} and \bt{}.
Empirically, improvements are marginal for 5 of the 6 datasets, only exceeding $1\%$ for \textsc{Bert-B} in a couple of instances where $N\le1000$.
\textsc{XLNet-B} and \textsc{RoBERTa-B} see no discernible improvements at almost any data level and just as frequently observe regressions in mean accuracy from \eda{} or \bt{}.
\mnli{} presents a clear outlier, with augmentations yielding relative improvements in excess of $2\%$, but only for \bert{}.
In contrast, the other pretrained transformers experience unpredictable, and mostly negative results.

In terms of augmentation preferences for \bert{}, \bt{} confers superior results to \eda{} in $60\%$ of cases, averaging $0.18\%$ absolute difference.
This advantage is muted for both \xlnet{} and \roberta{}, with only $53\%$ of cases preferring \bt{}, and at smaller margins.

Table~\ref{table:acc-improve} shows the improvement of \textbf{either} \eda{} or \bt{} over \textsc{No DA}, averaged across all 6 datasets.
We compare against \citet{wei2019eda}'s experiments, measuring the impact of \eda{} on \lstm{} and \textsc{CNN} models over 5 classification datasets.~\footnote{Their experimental setup is directly comparable to our own, comprising similar training sizes, datasets, and tuning procedures.}
They observe consistent improvements for non-pretrained models.
\lstm{}s and \textsc{CNN}s improve $3.8\%$ and $2.1\%$ on average at $N=500$ training points, and $0.9\%$ and $0.5\%$ on average with full data (approximately equivalent to our own \textsc{Full} setting).
As compared to these, \bert{} observes muted benefits.
To exclude \mnli{} from this average (not present in \citet{wei2019eda}'s experiments) would reduce all of \bert{}'s improvements well below $1\%$. 
\roberta{} and \xlnet{} again show no signs of improvement, frequently yielding worse results than the baseline, even with the best data augmentation.

Finally, we examine the claim that data augmentation confers an advantage with any statistical significance.
We use a one-sided t-test with null hypothesis that data augmentation confers a greater mean performance than without, using p-value of $.05$. 
Examining \bt{} and \eda{} vs \textsc{No DA} over all dataset and data sizes we reject the null hypothesis in $43\%$, $85\%$, and $87\%$ of cases for \bert{}, \xlnet{}, and \roberta{} respectively.
Moreover, for \roberta{}, the inverse hypothesis (that \textsc{No DA} is statistically better than DA) is true in $28\%$ of cases.

We believe these results are surprising due to two advantages given to data augmentation in this experimental setup: (A) these are relatively low data regimes compared to what is available for most tasks, and (B) in total, the data augmentation techniques receive twice the number of tuning trials as the \textsc{No DA} baseline.
Even if \eda{} and \bt{} confer no advantage over \textsc{No DA}, we would expect to see a minor positive increase from tuning over twice as many trials.~\footnote{Per dataset metrics, with confidence intervals, are available in Appendix C.}

\begin{table}[t]
	\centering
	\resizebox{\linewidth}{!}{
			\begin{tabular}{lrrrrr}
				\toprule
				Model & \multicolumn{5}{c}{Train Size ($N$)} \\
				\midrule
                {} & 500 & 1000 & 2000 & 3000 & Full \\
                \midrule
                \textsc{LSTM $^\dagger$} & +3.8 & - & +0.7 & - & +0.9 \\
                \textsc{CNN $^\dagger$} & +2.1 & - & +0.8 & - & +0.5 \\
                \textsc{Bert-B} & +1.13 & +1.23 & +0.82 & +0.78 & +0.61 \\
                \textsc{XLNet-B} & +0.56 & +0.02 & +0.22 & -0.02 & -0.01 \\
                \textsc{RoBERTa-B} & -0.14 & -0.04 & +0.02 & +0.27 & +0.03 \\
				\bottomrule
			\end{tabular}}
	\caption{
        The absolute improvement in test accuracy (\%) by \textbf{either} data augmentation technique over \textsc{No DA}. Results are averaged over all 6 datasets. \\
        $^\dagger$ We include \citet{wei2019eda}'s results for comparison, though their setup differs slightly: they show the improvement only for \eda{} (not the best of \eda{} and \bt{}), and they average over 5 classification datasets, of which we have \sst{}, \subj{}, and \trec{} in common.}
	\label{table:acc-improve}
	\vspace*{-.4cm}
\end{table}

\respace
\section{Discussion \& Limitations}
\respace

Our empirical results verify that popular data augmentation techniques fail to consistently improve performance for modern pretrained transformers --- even when training data is limited.
A single exception (\textsc{Bert-B} on \mnli{}) sees significant benefits from data augmentation.
We speculate the outlier results could pertain to the inherent difficulty of natural language inference in low data regimes.
Alternatively, \citet{gururangan-etal-2018-annotation} discuss ``annotation artifacts" in \mnli{} that lead models to rely on simple heuristics, such as the presence of the word ``not" in order to make classifications.
\eda{} or \bt{} could mitigate these spurious cues by distributing artifacts more evenly across labels. 

\subsection{Why can Data Augmentation be ineffective?}

Our results consistently demonstrate that augmentation provides more benefits to \bert{} than to \roberta{} and \xlnet{}. 
The key distinguishing factor between these models is the scale of unsupervised pretraining; therefore, we hypothesize that pretraining provides the same benefits targeted by common augmentation techniques.
\citet{arora2020contextual} characterize the benefits of contextual embeddings, showing a boost on tasks containing complex linguistic structures, ambiguous word usage, and unseen words.
Text editing and translation techniques vary linguistic structure and word usage to address these same issues.
Under this hypothesis, we would expect new data augmentation techniques to help only when they provide linguistic patterns that are relevant to the task but not seen during pretraining.

Manually inspecting \rt{} examples for which an \lstm{} requires augmentation to classify correctly, but \roberta{} does not, we observe rare word choice, atypical sentence structure and generally off-beat reviews. 
This set contains reviews such as “suffers from over-familiarity since hit-hungry british filmmakers have strip-mined the monty formula mercilessly since 1997”, “wishy-washy”, or “wanker goths are on the loose! run for your lives!”, as compared to “exceptionally well acted by diane lane and richard gere”, more representative of examples outside this set. 
We verify this quantitatively: for 100 examples in this set there are 206 (rare) words which \textbf{only} appear in this set, whereas for $100$ random samples we see an average of $116$ rare words. 
Interestingly, we also notice label skew in this set ($34\%$ of examples are positive instead of the overall mean of $50\%$).
While we leave deeper analysis to future work, we believe these results suggest data augmentation and pretraining both improve a model's ability to handle complex linguistic structure, ambiguous word usage, and unseen words within a category of label.

\subsection{When can Data Augmentation be useful?}

Given these observations, where might task-agnostic data augmentation be useful (with pretrained models)? 
One candidate application is out-of-domain generalization.
However, we believe the target domain must \textbf{not} be well represented in the pretraining corpus.
For instance, \citet{longpre2019exploration} did not find standard \bt{} useful for improving generalization of question answering models.
While their training domains are diverse, they are mostly based in Wikipedia and other common sources well represented in the \bert{} pretraining corpus.
Additionally, we suspect it is not enough to vary/modify examples in ways already seen in pretraining.
Our results motivate more sophisticated (read: targeted) augmentation techniques rather than generic, task (and domain)-agnostic strategies which unsupervised pretraining may capture more effectively.

Another candidate application of task-agnostic data augmentation is semi-supervised learning. 
\citet{xie2019unsupervised} illustrate a use for generic data augmentations as a noising agent for their consistency training method, assuming large quantities of unlabeled, in-domain data are available.
While task-agnostic data augmentations are effective in this particular task setup, they are not the critical factor in the success of the method, nor is it clear that more tailored or alternative noising techniques might not achieve even greater success.

To our knowledge, our experiments provide the most extensive examination of task-agnostic data augmentation for pretrained transformers. 
Nonetheless, our scope has been limited to classification tasks, and to the more common models and augmentations techniques.


\respace
\section{Conclusion}
\respace

We examine the effect of task-agnostic data augmentation in modern pretrained transformers.
Isolating low data regimes ($<10k$ training data points) across a range of factors, we observe a negative result: popular augmentation techniques fail to consistently improve performance for modern pretrained transformers. 
Further, we provide empirical evidence that suggests the scale of pretraining may be the primary factor in the diminished efficacy of textual augmentations.
We hope our work provides guidance to ML practitioners in deciding when to use data augmentation and encourages further examination of its relationship to unsupervised pretraining.

\respace
\section{Acknowledgments} 
\respace
We would like to acknowledge Stephen Pulman, Andrew Fandrianto, Drew Frank, Leon Gatys, Silviana Ciurea-Ilcus, Hang Zhao, and Kanit Wongsuphasawat for their guiding insights and helpful discussion.

\bibliography{emnlp2020}
\bibliographystyle{acl_natbib}

\clearpage
\appendix

\label{sec:appendix}

\section{Reproducibility}

\subsection{Transformer Models and Training}

We share the details of our hyper-parameter selection, for easy reproducibility. 
For each of \bert{}, \xlnet{}, and \roberta{} we use configurations mostly consistent with their original releases' recommendations.

In all cases code is adapted with minimal changes from open source repositories. The majority of changes to each repository pertain to supporting all 6 datasets, their augmentations, as well as better metrics reporting.
All models were trained on $1$ NVIDIA Tesla V100 GPU.

For each model we tune over 4 hyperparameters to which the final performance was particularly sensitive. 
The ``augmentation dose" parameter, as described in the paper, only applies to models trained with either \eda{} or \bt{}.
We verify in Appendix Section B that the addition of this tuning dimension did not alter our conclusions with respect to the impact of data augmentation when fully tuned.
Lastly, we would note that the final model size varies slightly depending on the size of the classification head --- dictated by the number of classes in the task.

\subsection{Bert-Base}

For \bert{} \citep{devlin2019bert} we use the original implementation in TensorFlow \citep{tensorflow2015-whitepaper}.\footnote{Code adapted from \url{https://github.com/google-research/bert}.} 
See Table~\ref{appendix-bert-hyperparams} for details in our training setup and hyperparameter tuning ranges.

\subsection{XLNet-Base}

For \xlnet{} \citep{yang2019xlnet} we also use the original implementation in TensorFlow.\footnote{Code adapted from \url{https://github.com/zihangdai/xlnet}.}
See Table~\ref{appendix-xlnet-hyperparams} for details in our training setup and hyperparameter tuning ranges.

\subsection{RoBERTa-Base}

For \roberta{} \citep{liu2019roberta} we use a standard PyTorch \citep{NEURIPS2019_9015} implementation as provided by HuggingFace.\footnote{Code adapted from \url{https://github.com/huggingface/transformers}.} 
See Table~\ref{appendix-roberta-hyperparams} for details in our training setup and hyperparameter tuning ranges.

\begin{table}[htb]
\small
\centering
\begin{tabular}{ll}
\toprule
\textsc{Model Parameters} & \textsc{Value/Range} \\
\midrule
\textbf{Fixed Parameters} & {} \\
\midrule
Batch Size & 50 \\
Optimizer & Adam \\
Learning Rate Schedule & Exponential Decay\\
Lower Case & True \\
Max Sequence Length & 100 \\
\midrule
\textbf{Tuned Parameters} & {} \\
\midrule
Num Epochs & [$2$, $100$] \\
Dropout & [$0.05$, $0.15$] \\
Learning Rate & [$1e-5$, $5e-5$] \\
Augmentation Dose & [$0.5$, $2.0$] \\
\midrule
\textbf{Extra Info} & {} \\
\midrule
Model Size (\# params) & $108.3M$ \\
Vocab Size & $30,522$ \\
Avg. Runtime (Full data) & $46m$ \\
\bottomrule
\end{tabular}
\caption{\label{appendix-bert-hyperparams}
Hyperparameter selection and tuning ranges for \textsc{Bert-Base}.
}
\end{table}

\begin{table}[htb]
\small
\centering
\begin{tabular}{ll}
\toprule
\textsc{Model Parameters} & \textsc{Value/Range} \\
\midrule
\textbf{Fixed Parameters} & {} \\
\midrule
Batch Size & 12 \\
Optimizer & Adam \\
Learning Rate Schedule & Exponential Decay\\
Lower Case & True \\
Max Sequence Length & 100 \\
\midrule
\textbf{Tuned Parameters} & {} \\
\midrule
Num Epochs & [$2$, $20$] \\
Dropout & [$0.05$, $0.15$] \\
Learning Rate & [$1e-5$, $5e-5$] \\
Augmentation Dose & [$0.5$, $2.0$] \\
\midrule
\textbf{Extra Info} & {} \\
\midrule
Model Size (\# params) & $117.3M$ \\
Vocab Size & $32,000$ \\
Avg. Runtime (Full data) & $37m$ \\
\bottomrule
\end{tabular}
\caption{\label{appendix-xlnet-hyperparams}
Hyperparameter selection and tuning ranges for \textsc{XLnet-Base}.
}
\end{table}

\begin{table}[htb]
\small
\centering
\begin{tabular}{ll}
\toprule
\textsc{Model Parameters} & \textsc{Value/Range} \\
\midrule
\textbf{Fixed Parameters} & {} \\
\midrule
Batch Size & 50 \\
Optimizer & Adam \\
Learning Rate Schedule & Exponential Decay\\
Lower Case & False \\
Max Sequence Length & 100 \\
\midrule
\textbf{Tuned Parameters} & {} \\
\midrule
Num Epochs & [$2$, $20$] \\
Dropout & [$0.05$, $0.15$] \\
Learning Rate & [$1e-5$, $5e-5$] \\
Augmentation Dose & [$0.5$, $2.0$] \\
\midrule
\textbf{Extra Info} & {} \\
\midrule
Model Size (\# params) & $125.2M$ \\
Vocab Size & $50,265$ \\
Avg. Runtime (Full data) & $32m$ \\
\bottomrule
\end{tabular}
\caption{\label{appendix-roberta-hyperparams}
Hyperparameter selection and tuning ranges for \textsc{RoBERTa-Base}.
}
\end{table}

\subsection{Datasets}

We experiemnt with 6 classification datasets.
These are \sst{} \citep{socher2013recursive}\footnote{Available at \url{https://nlp.stanford.edu/sentiment/}} and \rt{} \citep{pang2005seeing}\footnote{Available at \url{http://www.cs.cornell.edu/people/pabo/movie-review-data/}} for sentiment analysis, \subj{} \citep{10.3115/1218955.1218990}\footnote{Available at \url{http://www.cs.cornell.edu/people/pabo/movie-review-data/}} for subjectivity detection, \trec{} \citep{10.3115/1072228.1072378}\footnote{Available at \url{https://cogcomp.seas.upenn.edu/Data/QA/QC/}} for question type classification, \sts{} \citep{baudivs2016sentence}\footnote{Available at \url{https://gluebenchmark.com/tasks}} for semantic similarity, and \mnli{} \citep{williams2017broad}\footnote{Available at \url{https://www.nyu.edu/projects/bowman/multinli/}} for natural language inference.
For each of these we randomly sample up to $10k$ data points (if available) from the training sets, and separate out $1k$ for each of validation and testing. Additional statistics are available in the main paper.

\section{Verifying Tuning Sufficiency}

To ensure our conclusions are reliable we must verify that our tuning is sufficient to capture all the benefits of data augmentation. 
Accordingly, we double the number of hyperparameter configurations ($K$) and see if any of the conclusions change.
As this experiment is computationally expensive, we benchmark the results only for \bert{} on \sst{}.
Full results are shown in Table~\ref{table:tune-check}.

\begin{table*}[]
\begin{tabular}{llrrrrr}
\toprule
\textsc{Dataset} & \textsc{D. Aug.} & \textsc{N=500} & \textsc{N=1000} & \textsc{N=2000} & \textsc{N=3000} & \textsc{Full} \\
\midrule\multirow{3}{*}{K$_A$ = 30}&\textsc{No DA.}&84.29 $\pm$ 0.02&85.50 $\pm$ 0.00&87.46 $\pm$ 0.01&88.40 $\pm$ 0.01&89.79 $\pm$ 0.00\\
 &\textsc{BT}&85.38 $\pm$ 0.01&87.35 $\pm$ 0.02&87.37 $\pm$ 0.02&89.29 $\pm$ 0.01&90.36 $\pm$ 0.00\\
 &\textsc{EDA}&85.06 $\pm$ 0.01&86.70 $\pm$ 0.02&87.83 $\pm$ 0.01&88.55 $\pm$ 0.01&90.57 $\pm$ 0.01\\
\midrule\multirow{3}{*}{K$_B$ = 60}&\textsc{No DA.}&84.64 $\pm$ 0.02&85.72 $\pm$ 0.01&87.80 $\pm$ 0.01&88.66 $\pm$ 0.01&89.77 $\pm$ 0.01\\
 &\textsc{BT}&85.30 $\pm$ 0.01&86.78 $\pm$ 0.01&87.69 $\pm$ 0.01&89.36 $\pm$ 0.01&90.25 $\pm$ 0.00\\
 &\textsc{EDA}&85.90 $\pm$ 0.01&87.22 $\pm$ 0.00&87.48 $\pm$ 0.01&88.52 $\pm$ 0.01&90.25 $\pm$ 0.01\\
\midrule

\multirow{3}{*}{K$_B$ - K$_A$}&\textsc{No DA.}& +0.38 & +0.08 & +0.03 & +0.11 & -0.06 \\
 &\textsc{BT}& +0.25 & -0.12 & +0.20 & +0.18 & +0.01 \\
 &\textsc{EDA}& +0.51 & +0.20 & +0.17 & +0.03 & -0.06 \\
 \midrule
\multicolumn{2}{c}{\textsc{Mean} (K$_B$ - K$_A$)}& +0.38 & +0.05 & +0.13 & +0.11 & -0.04\\
\midrule
\multicolumn{2}{c}{$\triangle(\text{DA} - \text{No DA})$}& +0.13 & -0.15 & +0.17 & +0.06 & +0.07\\
\bottomrule
\end{tabular}
\caption{Here we verify that $K=30$ hyperparameter trials is sufficient to accurately estimate the benefit of data augmentation with full tuning. For \textsc{Bert-B} on \sst{} we compare the results of $K_{A}=30$ (used in the paper) and $K_{B}=60$. \textsc{Mean} $K_{B} - K_{A}$ compares the average difference in mean performances by doubling the number of trials.  $\triangle(\text{DA} - \text{No DA})$ measures the difference between the accuracies of the best data augmentation technique and \textsc{No DA}.}
\label{table:tune-check}
\end{table*}

We observe that on average, doubling the number of configuration trials from $K_A=30$ to $K_B=60$ results in minor accuracy improvements at lower training set sizes (e.g. $+0.38$ at $N=500$), and negligible variations at higher training set sizes (e.g. $-0.04$ at $N=Full$).
We also measure the resulting change in the difference between using and not using any data augmentation ($\triangle(\text{DA} - \text{No DA})$).
While improvements are reported in favour of data augmentation over \textsc{No DA}, they are all $< 0.15\%$, indicating at $K=30$ our conclusions are robust.

\section{Empirical Results}

Detailed results are provided for analysis. 
In each results table we include the mean accuracy and $95\%$ confidence interval, for every dataset, augmentation type, and training data size.
These are the outputs of the second stage of tuning ``\textsc{SelectBest}" that use the best hyperparameters per setting in the first \textsc{RandomSearch} stage.
We select only the top $10$ trials of $20$ (by validation accuracy) to compute these test statistics, due to the observed volatility in fine-tuning Transformers with different seeds \citep{dodge2020fine}.

The full results are shown below for \textsc{Bert-Base} (Table~\ref{table:bbu}), for \textsc{XLNet-Base} (see Table~\ref{table:xlc}), and for \textsc{RoBERTa-Base} (see Table~\ref{table:rbc}).

\begin{table*}[]
\begin{tabular}{llrrrrr}
\toprule
\textsc{Dataset} & \textsc{D. Aug.} & \textsc{N=500} & \textsc{N=1000} & \textsc{N=2000} & \textsc{N=3000} & \textsc{Full} \\
\midrule\multirow{3}{*}{\rt{}}&\textsc{No DA.}&81.82 $\pm$ 0.02&83.23 $\pm$ 0.01&84.44 $\pm$ 0.01&85.08 $\pm$ 0.01&86.98 $\pm$ 0.00\\
 &\textsc{BT}&82.84 $\pm$ 0.01&83.43 $\pm$ 0.01&85.15 $\pm$ 0.01&85.64 $\pm$ 0.01&87.62 $\pm$ 0.01\\
 &\textsc{EDA}&81.26 $\pm$ 0.01&83.52 $\pm$ 0.01&85.17 $\pm$ 0.01&85.41 $\pm$ 0.01&87.57 $\pm$ 0.00\\
\midrule\multirow{3}{*}{\subj{}}&\textsc{No DA.}&94.07 $\pm$ 0.00&95.16 $\pm$ 0.01&96.34 $\pm$ 0.00&96.41 $\pm$ 0.01&97.40 $\pm$ 0.00\\
 &\textsc{BT}&95.09 $\pm$ 0.01&95.27 $\pm$ 0.01&96.28 $\pm$ 0.00&96.43 $\pm$ 0.00&98.04 $\pm$ 0.00\\
 &\textsc{EDA}&94.81 $\pm$ 0.01&95.68 $\pm$ 0.01&96.13 $\pm$ 0.00&96.77 $\pm$ 0.00&98.05 $\pm$ 0.00\\
\midrule\multirow{3}{*}{\sst{}}&\textsc{No DA.}&84.29 $\pm$ 0.02&85.50 $\pm$ 0.00&87.46 $\pm$ 0.01&88.40 $\pm$ 0.01&89.79 $\pm$ 0.00\\
 &\textsc{BT}&85.38 $\pm$ 0.01&87.35 $\pm$ 0.02&87.37 $\pm$ 0.02&89.29 $\pm$ 0.01&90.36 $\pm$ 0.00\\
 &\textsc{EDA}&85.06 $\pm$ 0.01&86.70 $\pm$ 0.02&87.83 $\pm$ 0.01&88.55 $\pm$ 0.01&90.57 $\pm$ 0.01\\
\midrule\multirow{3}{*}{\trec{}}&\textsc{No DA.}&87.51 $\pm$ 0.01&91.25 $\pm$ 0.01&93.00 $\pm$ 0.00&94.23 $\pm$ 0.00&94.67 $\pm$ 0.00\\
 &\textsc{BT}&88.95 $\pm$ 0.01&91.24 $\pm$ 0.01&93.42 $\pm$ 0.01&94.31 $\pm$ 0.00&94.59 $\pm$ 0.00\\
 &\textsc{EDA}&88.75 $\pm$ 0.01&92.25 $\pm$ 0.01&93.76 $\pm$ 0.00&94.31 $\pm$ 0.00&94.51 $\pm$ 0.00\\
\midrule\multirow{3}{*}{\mnli{}}&\textsc{No DA.}&47.29 $\pm$ 0.04&54.18 $\pm$ 0.07&62.50 $\pm$ 0.03&65.92 $\pm$ 0.03&72.90 $\pm$ 0.01\\
 &\textsc{BT}&49.23 $\pm$ 0.02&58.15 $\pm$ 0.03&66.46 $\pm$ 0.03&68.84 $\pm$ 0.03&74.25 $\pm$ 0.02\\
 &\textsc{EDA}&50.03 $\pm$ 0.03&56.92 $\pm$ 0.03&64.88 $\pm$ 0.02&67.04 $\pm$ 0.03&73.85 $\pm$ 0.02\\
\midrule\multirow{3}{*}{\sts{}}&\textsc{No DA.}&77.93 $\pm$ 0.04&84.61 $\pm$ 0.01&87.26 $\pm$ 0.00&87.73 $\pm$ 0.01&88.40 $\pm$ 0.01\\
 &\textsc{BT}&77.94 $\pm$ 0.05&84.97 $\pm$ 0.01&86.62 $\pm$ 0.01&88.13 $\pm$ 0.01&88.35 $\pm$ 0.01\\
 &\textsc{EDA}&78.27 $\pm$ 0.06&84.83 $\pm$ 0.02&86.89 $\pm$ 0.00&88.09 $\pm$ 0.00&88.56 $\pm$ 0.00\\
\bottomrule
\end{tabular}
\caption{\textsc{Bert-Base} mean test accuracy and the $95\%$ confidence interval for each task, augmentation, and data size, computed over the top $10$ best trials, by validation score.}
\label{table:bbu}
\end{table*}

\begin{table*}[]
\begin{tabular}{llrrrrr} 	
\toprule
\textsc{Dataset} & \textsc{D. Aug.} & \textsc{N=500} & \textsc{N=1000} & \textsc{N=2000} & \textsc{N=3000} & \textsc{Full} \\
\midrule\multirow{3}{*}{\rt{}}&\textsc{No DA.}&83.85 $\pm$ 0.01&86.58 $\pm$ 0.01&87.69 $\pm$ 0.01&88.59 $\pm$ 0.00&89.97 $\pm$ 0.01\\
 &\textsc{BT}&86.34 $\pm$ 0.01&86.77 $\pm$ 0.01&88.23 $\pm$ 0.00&88.80 $\pm$ 0.00&89.97 $\pm$ 0.01\\
 &\textsc{EDA}&84.71 $\pm$ 0.01&86.68 $\pm$ 0.01&87.54 $\pm$ 0.00&88.61 $\pm$ 0.00&89.94 $\pm$ 0.00\\
\midrule\multirow{3}{*}{\subj{}}&\textsc{No DA.}&94.88 $\pm$ 0.00&95.65 $\pm$ 0.00&95.99 $\pm$ 0.00&96.29 $\pm$ 0.00&97.28 $\pm$ 0.00\\
 &\textsc{BT}&95.23 $\pm$ 0.00&96.07 $\pm$ 0.00&96.52 $\pm$ 0.00&96.65 $\pm$ 0.00&97.40 $\pm$ 0.00\\
 &\textsc{EDA}&94.69 $\pm$ 0.01&95.75 $\pm$ 0.01&96.41 $\pm$ 0.00&96.62 $\pm$ 0.00&97.5 $\pm$ 0.00\\
\midrule\multirow{3}{*}{\sst{}}&\textsc{No DA.}&89.44 $\pm$ 0.01&90.10 $\pm$ 0.00&91.20 $\pm$ 0.01&91.87 $\pm$ 0.01&92.98 $\pm$ 0.00\\
 &\textsc{BT}&89.43 $\pm$ 0.01&90.36 $\pm$ 0.01&91.39 $\pm$ 0.00&92.00 $\pm$ 0.00&92.88 $\pm$ 0.00\\
 &\textsc{EDA}&89.07 $\pm$ 0.01&90.45 $\pm$ 0.00&91.59 $\pm$ 0.00&91.49 $\pm$ 0.00&92.5 $\pm$ 0.00\\
\midrule\multirow{3}{*}{\trec{}}&\textsc{No DA.}&90.36 $\pm$ 0.00&92.46 $\pm$ 0.01&93.94 $\pm$ 0.00&95.07 $\pm$ 0.00&94.85 $\pm$ 0.00\\
 &\textsc{BT}&90.03 $\pm$ 0.01&92.16 $\pm$ 0.00&93.14 $\pm$ 0.00&94.56 $\pm$ 0.00&94.74 $\pm$ 0.00\\
 &\textsc{EDA}&90.11 $\pm$ 0.01&92.51 $\pm$ 0.00&94.14 $\pm$ 0.00&94.64 $\pm$ 0.00&95.16 $\pm$ 0.00\\
\midrule\multirow{3}{*}{\mnli{}}&\textsc{No DA.}&57.32 $\pm$ 0.02&65.80 $\pm$ 0.01&72.07 $\pm$ 0.01&74.97 $\pm$ 0.01&78.75 $\pm$ 0.01\\
 &\textsc{BT}&58.88 $\pm$ 0.02&65.49 $\pm$ 0.02&71.67 $\pm$ 0.01&74.16 $\pm$ 0.02&79.08 $\pm$ 0.01\\
 &\textsc{EDA}&56.90 $\pm$ 0.02&64.65 $\pm$ 0.02&72.48 $\pm$ 0.01&74.81 $\pm$ 0.00&78.87 $\pm$ 0.01\\
\midrule\multirow{3}{*}{\sts{}}&\textsc{No DA.}&73.76 $\pm$ 0.02&82.88 $\pm$ 0.00&86.29 $\pm$ 0.00&87.52 $\pm$ 0.00&88.35 $\pm$ 0.00\\
 &\textsc{BT}&74.82 $\pm$ 0.01&82.17 $\pm$ 0.01&85.73 $\pm$ 0.00&86.96 $\pm$ 0.00&87.52 $\pm$ 0.00\\
 &\textsc{EDA}&73.18 $\pm$ 0.01&82.87 $\pm$ 0.00&85.90 $\pm$ 0.00&87.05 $\pm$ 0.00&87.98 $\pm$ 0.00\\
\bottomrule
\end{tabular}
\caption{\textsc{XLNet-Base} mean test accuracy and the $95\%$ confidence interval for each task, augmentation, and data size, computed over the top $10$ best trials, by validation score.}
\label{table:xlc}
\end{table*}

\begin{table*}[]
\begin{tabular}{llrrrrr} 
\toprule
\textsc{Dataset} & \textsc{D. Aug.} & \textsc{N=500} & \textsc{N=1000} & \textsc{N=2000} & \textsc{N=3000} & \textsc{Full} \\
\midrule\multirow{3}{*}{\rt{}}&\textsc{No DA.}&84.84 $\pm$ 0.01&86.71 $\pm$ 0.00&87.05 $\pm$ 0.01&87.99 $\pm$ 0.01&90.10 $\pm$ 0.00\\
 &\textsc{BT}&84.66 $\pm$ 0.01&86.00 $\pm$ 0.01&87.48 $\pm$ 0.01&88.44 $\pm$ 0.01&90.08 $\pm$ 0.00\\
 &\textsc{EDA}&84.26 $\pm$ 0.01&86.53 $\pm$ 0.01&87.89 $\pm$ 0.01&88.40 $\pm$ 0.00&90.19 $\pm$ 0.01\\
\midrule\multirow{3}{*}{\subj{}}&\textsc{No DA.}&94.27 $\pm$ 0.00&95.50 $\pm$ 0.01&96.22 $\pm$ 0.00&96.48 $\pm$ 0.00&97.36 $\pm$ 0.00\\
 &\textsc{BT}&95.14 $\pm$ 0.00&95.74 $\pm$ 0.00&96.11 $\pm$ 0.00&96.28 $\pm$ 0.00&97.05 $\pm$ 0.00\\
 &\textsc{EDA}&94.55 $\pm$ 0.00&95.42 $\pm$ 0.01&95.87 $\pm$ 0.00&96.50 $\pm$ 0.00&97.31 $\pm$ 0.00\\
\midrule\multirow{3}{*}{\sst{}}&\textsc{No DA.}&90.80 $\pm$ 0.00&91.51 $\pm$ 0.00&91.95 $\pm$ 0.01&92.28 $\pm$ 0.01&93.52 $\pm$ 0.00\\
 &\textsc{BT}&90.13 $\pm$ 0.01&91.64 $\pm$ 0.00&92.75 $\pm$ 0.00&92.45 $\pm$ 0.00&93.96 $\pm$ 0.00\\
 &\textsc{EDA}&90.16 $\pm$ 0.01&91.58 $\pm$ 0.00&92.55 $\pm$ 0.00&92.92 $\pm$ 0.01&93.85 $\pm$ 0.00\\
\midrule\multirow{3}{*}{\trec{}}&\textsc{No DA.}&90.77 $\pm$ 0.00&93.41 $\pm$ 0.00&94.80 $\pm$ 0.01&95.03 $\pm$ 0.01&95.66 $\pm$ 0.00\\
 &\textsc{BT}&90.65 $\pm$ 0.00&93.04 $\pm$ 0.00&94.66 $\pm$ 0.00&94.99 $\pm$ 0.00&95.46 $\pm$ 0.00\\
 &\textsc{EDA}&90.97 $\pm$ 0.01&93.44 $\pm$ 0.00&94.70 $\pm$ 0.00&94.80 $\pm$ 0.00&95.14 $\pm$ 0.00\\
\midrule\multirow{3}{*}{\mnli{}}&\textsc{No DA.}&63.3 $\pm$ 0.04&73.18 $\pm$ 0.02&77.94 $\pm$ 0.01&77.69 $\pm$ 0.07&83.04 $\pm$ 0.00\\
 &\textsc{BT}&60.9 $\pm$ 0.17&72.04 $\pm$ 0.03&77.42 $\pm$ 0.01&79.14 $\pm$ 0.01&82.28 $\pm$ 0.01\\
 &\textsc{EDA}&61.15 $\pm$ 0.08&71.09 $\pm$ 0.03&72.59 $\pm$ 0.23&78.6 $\pm$ 0.02&83.5 $\pm$ 0.00\\
\midrule\multirow{3}{*}{\sts{}}&\textsc{No DA.}&79.49 $\pm$ 0.02&85.77 $\pm$ 0.01&89.32 $\pm$ 0.00&89.94 $\pm$ 0.00&90.29 $\pm$ 0.00\\
 &\textsc{BT}&79.24 $\pm$ 0.01&85.76 $\pm$ 0.01&88.80 $\pm$ 0.00&89.80 $\pm$ 0.00&89.79 $\pm$ 0.00\\
 &\textsc{EDA}&78.87 $\pm$ 0.01&84.95 $\pm$ 0.02&88.82 $\pm$ 0.00&89.92 $\pm$ 0.00&90.15 $\pm$ 0.00\\
\bottomrule
\end{tabular}
\caption{\textsc{RoBERTa-Base} mean test accuracy and the $95\%$ confidence interval for each task, augmentation, and data size, computed over the top $10$ best trials, by validation score.}
\label{table:rbc}
\end{table*}

\end{document}